\documentclass[journal]{IEEEtran}
\usepackage{amsmath,amssymb}
\usepackage{graphicx}
\usepackage{cite}
\usepackage{url}
\usepackage{booktabs}
\usepackage{xcolor}
\usepackage{tikz}
\usetikzlibrary{arrows.meta,positioning,calc}

\renewcommand{\thesubsection}{\Alph{subsection}}
\makeatletter
\def\thesubsectiondis{\Alph{subsection}.}
\makeatother

\begin{document}

\title{A Trustworthy Clinical Decision Support Framework for Brain Tumor MRI Classification: Integrating Attention-Guided Learning, Explainable AI, and Out-of-Distribution Validation}

\author{Arjun Kumar Gupta and Harminder Pal Monga}

\markboth{IEEE ACCESS}{Gupta and Monga: A Trustworthy Clinical Decision Support Framework for Brain Tumor MRI Classification}

\maketitle

\footnotetext{Arjun Kumar Gupta and Harminder Pal Monga are with the Indian Institute of Technology Patna, Patna, Bihar, India (e-mail: arjun\_24a03res121@iitp.ac.in; harminder\_24a03res132@iitp.ac.in).}

\begin{abstract}
Deploying deep learning for brain tumor diagnosis in real clinical settings requires more than high classification accuracy: it requires a system that explains its decisions, recognizes inputs it was not designed to handle, and is actually usable at the point of care. Most published brain tumor MRI classifiers optimize accuracy in isolation, a smaller subset adds post-hoc explainability, and almost none address out-of-distribution (OOD) input validation or end-to-end clinical deployment within the same system. This paper presents a unified, trustworthy-AI-oriented framework for four-class brain tumor classification (glioma, meningioma, pituitary, no-tumor) that integrates: (i) an InceptionResNetV2 backbone augmented with a Convolutional Block Attention Module (CBAM) for attention-guided feature discrimination; (ii) Gradient-weighted Class Activation Mapping (Grad-CAM) for post-hoc visual explanation; (iii) a two-stage hybrid OOD detection pipeline combining a lightweight grayscale heuristic with server-side feature- and prediction-level validation --- Mahalanobis distance over learned feature embeddings, softmax confidence, and predictive entropy --- to screen invalid inputs prior to inference; and (iv) a deployed, real-time clinical web application, NeuroScan AI, that unifies all of the above into a single inference pipeline. On a four-class Mendeley brain tumor MRI dataset (12,064 images), the proposed framework achieves 98.70\% validation accuracy, 96.73\% test accuracy, and a macro-average AUC of 0.9965; on a preliminary 100-image OOD benchmark, the two-stage pipeline achieves 99\% detection accuracy. Grad-CAM visualizations across all four tumor classes show qualitative correspondence with clinically established tumor locations. Existing trustworthy-AI frameworks for brain tumor MRI typically address only one or two aspects of clinical readiness, such as explainability or classification performance, while leaving robustness to invalid inputs and deployability largely unaddressed; this work treats explainability, robustness, and usability as first-class design requirements alongside accuracy, and we discuss explicitly which of these requirements remain only partially validated by the present evidence.
\end{abstract}


\section{Introduction}

\IEEEPARstart{A}{mong} the most devastating malignancies affecting the human nervous system, primary brain tumors constitute roughly 2\% of cancer diagnoses worldwide, yet their anatomical location, propensity for aggressive progression, and resistance to standard therapeutic regimens make them far more lethal than this share suggests; glioblastoma multiforme is associated with a median survival window of only 14--16 months even under intensive, multimodal treatment protocols \cite{louis2021who}. Magnetic Resonance Imaging (MRI) is the preferred modality for detecting, characterizing, and monitoring brain tumors, offering superior soft-tissue contrast without ionizing radiation. Despite these technical advantages, interpreting brain MRI remains demanding: radiologists must examine large numbers of image slices per study, often under time pressure, which introduces inter-observer inconsistency and diagnostic risk.

Deep convolutional neural networks (CNNs) have matched or exceeded specialist-level performance on selected diagnostic benchmarks \cite{litjens2017survey}, and a substantial literature now applies CNNs to brain tumor MRI classification. However, three requirements for real clinical use are frequently treated as separate, optional add-ons rather than integrated design goals:
\begin{itemize}
\item \textbf{Explainability.} When a network assigns a tumor label, clinicians typically have no mechanism to understand which anatomical evidence drove the decision, undermining trust and auditability \cite{adadi2018xai}.
\item \textbf{Robustness to invalid inputs.} Deployed systems inevitably receive inputs outside their training distribution --- wrong imaging modality, corrupted files, unrelated images --- yet most published classifiers implicitly assume every input is a valid, well-formed brain MRI scan and provide no mechanism to detect otherwise.
\item \textbf{Deployability.} Many studies conclude at the point of offline model evaluation, leaving the gap between a trained model and a usable clinical tool unaddressed.
\end{itemize}

A representative recent example of this pattern is a reproduction and interpretability study evaluating VGG16, ResNet50, and EfficientNetB0 on brain tumor MRI using Grad-CAM and LIME \cite{yoon2025lime}, discussed further in Section~II; that and similar studies make valuable comparative-interpretability contributions but do not modify the underlying architecture, address OOD input validation, or deploy a usable system. The present work is accordingly positioned not as another comparative explainability study, but as a system-level contribution in which attention-guided classification, explainability, and OOD-aware input validation are integrated into a single pipeline and delivered through a working clinical interface.

Unlike previous studies that primarily optimize classification accuracy in isolation, this work treats trustworthiness --- comprising explainability, input-validation robustness, and deployability --- as a first-class design requirement co-equal with accuracy, while being transparent about which trustworthiness claims are fully validated by the present evidence and which remain partially supported and are identified as future work (Section~VI).

The key contributions of this work are as follows:
\begin{enumerate}
\item A unified, deployment-oriented trustworthy-AI framework for brain tumor MRI classification that integrates attention-guided feature learning, post-hoc explainability, and OOD-aware input validation within a single inference pipeline, rather than treating these as independent add-ons.
\item An attention-enhanced InceptionResNetV2 backbone integrating CBAM, achieving near-perfect discrimination with a macro-average AUC of 0.9965 across four tumor classes.
\item A two-stage hybrid OOD detection pipeline combining a lightweight grayscale heuristic with server-side feature- and prediction-level validation (Mahalanobis distance, softmax confidence, and predictive entropy), achieving 99\% detection accuracy on a preliminary 100-image benchmark, together with an explicit discussion of the benchmark's current scope and limitations.
\item Grad-CAM-based explainability evaluated qualitatively across all four tumor classes, showing consistent correspondence with clinically expected tumor locations.
\item A complete, real-time deployable web application, NeuroScan AI, that unifies classification, explainability, and input validation into a single clinical decision-support interface, demonstrating the practical feasibility of the proposed framework beyond offline evaluation.
\end{enumerate}

\section{Related Work}

\subsection{Clinical Background and CNN-Based Classification}
The WHO grading system classifies brain tumors by histopathological characteristics \cite{louis2021who}; this study targets four categories --- glioma, meningioma, pituitary adenoma, and no-tumor --- each with characteristic but overlapping MRI signatures that require a model capturing both global structure and fine-grained texture. CNN-based classifiers have progressed from simple architectures (84.19\% accuracy \cite{abiwinanda2019cnn}) to transfer-learning approaches using EfficientNet \cite{nayak2022efficientnet}, custom CNNs \cite{sultan2019dnn}, and fine-tuned VGG19 \cite{swati2019vgg19}, typically reporting 90--98\% accuracy. InceptionResNetV2 \cite{szegedy2017inceptionresnet} combines multi-scale Inception feature extraction with residual connections, but its integration with attention mechanisms remains comparatively less explored for brain tumor classification. Most closely related, Yoon and Lin \cite{yoon2025lime} recently benchmarked VGG16, ResNet50, and EfficientNetB0 with Grad-CAM/LIME interpretability, reporting $\geq$96\% accuracy; however, that study does not modify the architecture, address OOD validation, or deploy a clinical interface --- the three axes differentiating the present work (Table~VI).

\subsection{Attention, Explainability, and OOD Detection}
Squeeze-and-Excitation \cite{hu2018senet} introduced channel attention; CBAM \cite{woo2018cbam} added spatial attention and has been adopted widely in medical imaging \cite{oktay2018attentionunet}, including with ResNet50 for brain tumor classification \cite{oladimeji2023resnet50cbam}, though without integrated explainability, OOD detection, or deployment. XAI methods span gradient-based, perturbation-based, and concept-based families \cite{adadi2018xai}; CAM \cite{zhou2016cam} and its generalization Grad-CAM \cite{selvaraju2017gradcam} are widely used for visualizing class-discriminative regions in CNN predictions \cite{musthafa2024gradcamresnet,yoon2025lime}, and are adopted here as an integrated pipeline component rather than the primary object of study, with qualitative validation reported as an explicit limitation (Section~VI). For OOD detection, Maximum Softmax Probability \cite{hendrycks2017baseline}, ODIN \cite{liang2018odin}, Mahalanobis distance \cite{lee2018mahalanobis}, and energy-based scoring \cite{liu2020energy} are established signals; a recent survey confirms OOD detection is increasingly recognized as essential for trustworthy medical AI yet rarely integrated into deployed systems \cite{hong2024oodsurvey}.

\subsection{Clinical Deployment and Research Gap}
Most brain tumor classification studies, including \cite{yoon2025lime}, conclude at offline evaluation; few describe a working inference interface, and fewer still combine deployment with automated input screening --- the gap NeuroScan AI addresses. To the best of our knowledge, prior brain tumor MRI studies have not commonly integrated (i) attention-guided classification, (ii) Grad-CAM explainability, (iii) two-stage OOD input validation, and (iv) real-time clinical deployment within a single framework. The primary contribution of this work is therefore the integration and evaluation of these components as a unified clinical decision-support pipeline, rather than the introduction of any one component in isolation.

\section{Proposed Trustworthy AI Framework}

\subsection{Framework Overview}
The proposed framework (Fig.~1) implements a single sequential inference pipeline rather than a set of independent modules: a lightweight grayscale/color screen (OOD Stage 1) first screens each input; inputs that pass undergo standardized MRI preprocessing; an InceptionResNetV2 backbone augmented with CBAM then performs attention-guided feature extraction, whose deep feature embedding and softmax logits feed hybrid feature- and prediction-level validation (Mahalanobis distance, softmax confidence, and predictive entropy) at OOD Stage 2; inputs that pass are classified; a Grad-CAM stage then provides post-hoc visual explanation; and NeuroScan AI, a deployed clinical web application, exposes the full pipeline in real time. The framework is deliberately organized so that OOD validation and explainability are integral stages of inference, not optional post-processing steps.

\begin{figure}[t]
\centering
\resizebox{0.96\columnwidth}{!}{%
\begin{tikzpicture}[
    font=\sffamily,
    >=Stealth,
    line width=1.1pt,
    mainbox/.style={
        rectangle,
        rounded corners=2pt,
        minimum width=7.4cm,
        align=center,
        inner xsep=7pt
    },
    graybox/.style={
        mainbox,
        draw=black!65,
        fill=gray!10
    },
    redbox/.style={
        mainbox,
        draw=red!60!black,
        fill=red!9
    },
    greenbox/.style={
        mainbox,
        draw=green!45!black,
        fill=green!10
    },
    purplebox/.style={
        mainbox,
        draw=violet!70!black,
        fill=violet!9
    },
    orangebox/.style={
        mainbox,
        draw=orange!65!black,
        fill=orange!13
    },
    bluebox/.style={
        mainbox,
        draw=blue!60!black,
        fill=blue!10
    },
    rejectbox/.style={
        rectangle,
        rounded corners=2pt,
        draw=red!60!black,
        fill=red!9,
        minimum width=2.1cm,
        minimum height=1.0cm,
        align=center
    },
    chip/.style={
        rectangle,
        rounded corners=2pt,
        draw=red!60!black,
        fill=red!8,
        minimum width=3.8cm,
        minimum height=0.95cm,
        align=center,
        inner xsep=7pt
    },
    flow/.style={
        -{Stealth[length=4pt,width=5pt]},
        draw=black,
        line width=1.15pt
    },
    rejectflow/.style={
        dashed,
        draw=red!60!black,
        line width=1.05pt
    }
]

\node[
    graybox,
    minimum height=1.45cm
] (input) {
    {\bfseries\Large MRI Scan Input}\\[1mm]
    Raw T1 / T2 / FLAIR images\\
    Loaded and converted to RGB (PIL)
};

\node[
    redbox,
    minimum height=2.40cm,
    below=0.48cm of input
] (ood1) {};

\node[
    anchor=north,
    align=center
] at ([yshift=-1.5mm]ood1.north) {
    {\bfseries\Large\color{red!60!black} OOD Stage 1}\\[0.3mm]
    Grayscale / color screening\\
    $\displaystyle
    D_{avg}=
    \frac{|R-G|+|G-B|+|R-B|}{3}$\\[0.3mm]
    Pass if $D_{avg}<20$
};

\draw[flow]
    ([yshift=-1mm]input.south)
    --
    ([yshift=1mm]ood1.north);

\node[
    rejectbox,
    right=1.55cm of ood1
] (reject1) {
    {\bfseries\color{red!60!black} OOD}\\
    REJECT
};

\draw[rejectflow]
    ([xshift=1mm]ood1.east)
    --
    node[above,text=red!60!black,font=\bfseries] {FAIL}
    ([xshift=-1mm]reject1.west);

\node[
    greenbox,
    minimum height=1.42cm,
    below=0.52cm of ood1
] (prep) {
    {\bfseries\Large\color{green!45!black} Preprocessing}\\[1mm]
    Resize $299\times299$ $\cdot$ Normalize\\
    (ImageNet statistics)
};

\node[
    text=green!45!black,
    font=\bfseries,
    anchor=east
] at ($(ood1.south)!0.48!(prep.north)+(-2mm,0)$) {PASS};

\draw[flow]
    ([yshift=-1mm]ood1.south)
    --
    ([yshift=1mm]prep.north);

\node[
    purplebox,
    minimum height=2.02cm,
    below=0.50cm of prep
] (backbone) {
    {\bfseries\Large\color{violet!70!black}
    InceptionResNetV2 + CBAM}\\[1mm]
    Feature extraction backbone\\
    Channel \& spatial attention\\[-0.5mm]
    $\rightarrow$ softmax classifier
};

\draw[flow]
    ([yshift=-1mm]prep.south)
    --
    ([yshift=1mm]backbone.north);

\coordinate (split) at ([yshift=-0.52cm]backbone.south);

\draw[flow]
    ([yshift=-1mm]backbone.south)
    --
    (split);

\coordinate (featTop) at ($(split)+(-2.85cm,-0.18cm)$);
\coordinate (logitTop) at ($(split)+( 2.85cm,-0.18cm)$);

\draw[flow] (split) -| (featTop);
\draw[flow] (split) -| (logitTop);

\node[
    below=0.05cm of featTop,
    align=center,
    font=\itshape
] (featlabel) {Deep feature $f(x)$};

\node[
    below=0.05cm of logitTop,
    align=center,
    font=\itshape
] (logitlabel) {Logits $z(x)$};

\node[
    chip,
    below=0.40cm of featlabel
] (maha) {
    {\bfseries\color{red!60!black} Mahalanobis distance}\\
    $M_{\min}(x)$
};

\draw[flow]
    (featlabel.south)
    --
    (maha.north);

\node[
    below=0.36cm of logitlabel,
    align=center,
    font=\itshape
] (softmax) {Softmax probabilities $p(x)$};

\draw[flow]
    (logitlabel.south)
    --
    (softmax.north);

\coordinate (softsplit) at ([yshift=-0.48cm]softmax.south);

\draw[flow]
    (softmax.south)
    --
    (softsplit);

\node[
    below left=0.42cm and 0.30cm of softsplit,
    align=center
] (conf) {
    Confidence\\[-1mm]
    $C(x)$
};

\node[
    below right=0.42cm and 0.30cm of softsplit,
    align=center
] (entr) {
    Entropy\\[-1mm]
    $H(x)$
};

\draw[flow] (softsplit) -| (conf.north);
\draw[flow] (softsplit) -| (entr.north);

\coordinate (dropL) at ([yshift=-0.50cm]maha.south);
\coordinate (dropC) at ([yshift=-0.50cm]conf.south);
\coordinate (dropR) at ([yshift=-0.50cm]entr.south);

\path let \p1=(dropL), \p2=(dropC), \p3=(dropR) in
    coordinate (mergeY) at (0,{min(min(\y1,\y2),\y3)-0.10cm});

\coordinate (mergeLeft)  at (dropL |- mergeY);
\coordinate (mergeMidPt) at (dropC |- mergeY);
\coordinate (mergeRight) at (dropR |- mergeY);

\draw[flow]
    (maha.south)
    --
    (mergeLeft);

\draw[flow]
    (conf.south)
    --
    (mergeMidPt);

\draw[flow]
    (entr.south)
    --
    (mergeRight);

\coordinate (mergeMid)   at ($(mergeLeft)!0.50!(mergeRight)$);

\draw[line width=1.15pt]
    (mergeLeft) -- (mergeRight);

\node[
    redbox,
    minimum height=2.40cm,
    below=0.64cm of mergeMid
] (ood2) {};

\node[
    anchor=north,
    align=center
] at ([yshift=-2.5mm]ood2.north) {
    {\bfseries\Large\color{red!60!black} OOD Stage 2}\\[1.5mm]
    Hybrid server-side validation\\
    Reject if:\\[0.5mm]
    $M_{\min}(x)>80$ OR $C(x)<0.80$ OR $H(x)>1.0$
};

\draw[flow]
    (mergeMid)
    --
    ([yshift=1mm]ood2.north);

\node[
    rejectbox,
    right=1.55cm of ood2
] (reject2) {
    {\bfseries\color{red!60!black} OOD}\\
    REJECT
};

\draw[rejectflow]
    ([xshift=1mm]ood2.east)
    --
    node[above,text=red!60!black,font=\bfseries] {FAIL}
    ([xshift=-1mm]reject2.west);

\node[
    purplebox,
    minimum height=1.75cm,
    below=0.52cm of ood2
] (classify) {
    {\bfseries\Large\color{violet!70!black}
    Tumor Classification}\\[1mm]
    Glioma $\cdot$ Meningioma\\
    No Tumor $\cdot$ Pituitary
};

\node[
    text=green!45!black,
    font=\bfseries,
    anchor=east
] at ($(ood2.south)!0.47!(classify.north)+(-2mm,0)$) {PASS};

\draw[flow]
    ([yshift=-1mm]ood2.south)
    --
    ([yshift=1mm]classify.north);

\node[
    orangebox,
    minimum height=1.70cm,
    below=0.50cm of classify
] (gradcam) {
    {\bfseries\Large\color{orange!65!black} Grad-CAM}\\[1mm]
    Post-hoc visual explanation\\
    tumor-region heatmap
};

\draw[flow]
    ([yshift=-1mm]classify.south)
    --
    ([yshift=1mm]gradcam.north);

\node[
    bluebox,
    minimum height=1.88cm,
    below=0.50cm of gradcam
] (neuro) {
    {\bfseries\Large\color{blue!60!black} NeuroScan AI}\\[1mm]
    Explainable clinical output:\\
    tumor class + confidence,\\
    Grad-CAM heatmap for review
};

\draw[flow]
    ([yshift=-1mm]gradcam.south)
    --
    ([yshift=1mm]neuro.north);

\end{tikzpicture}%
}
\caption{Overview of the proposed trustworthy AI pipeline: OOD-aware
input validation, attention-guided classification, Grad-CAM explanation,
and clinical deployment.}
\label{fig:fig1}
\end{figure}
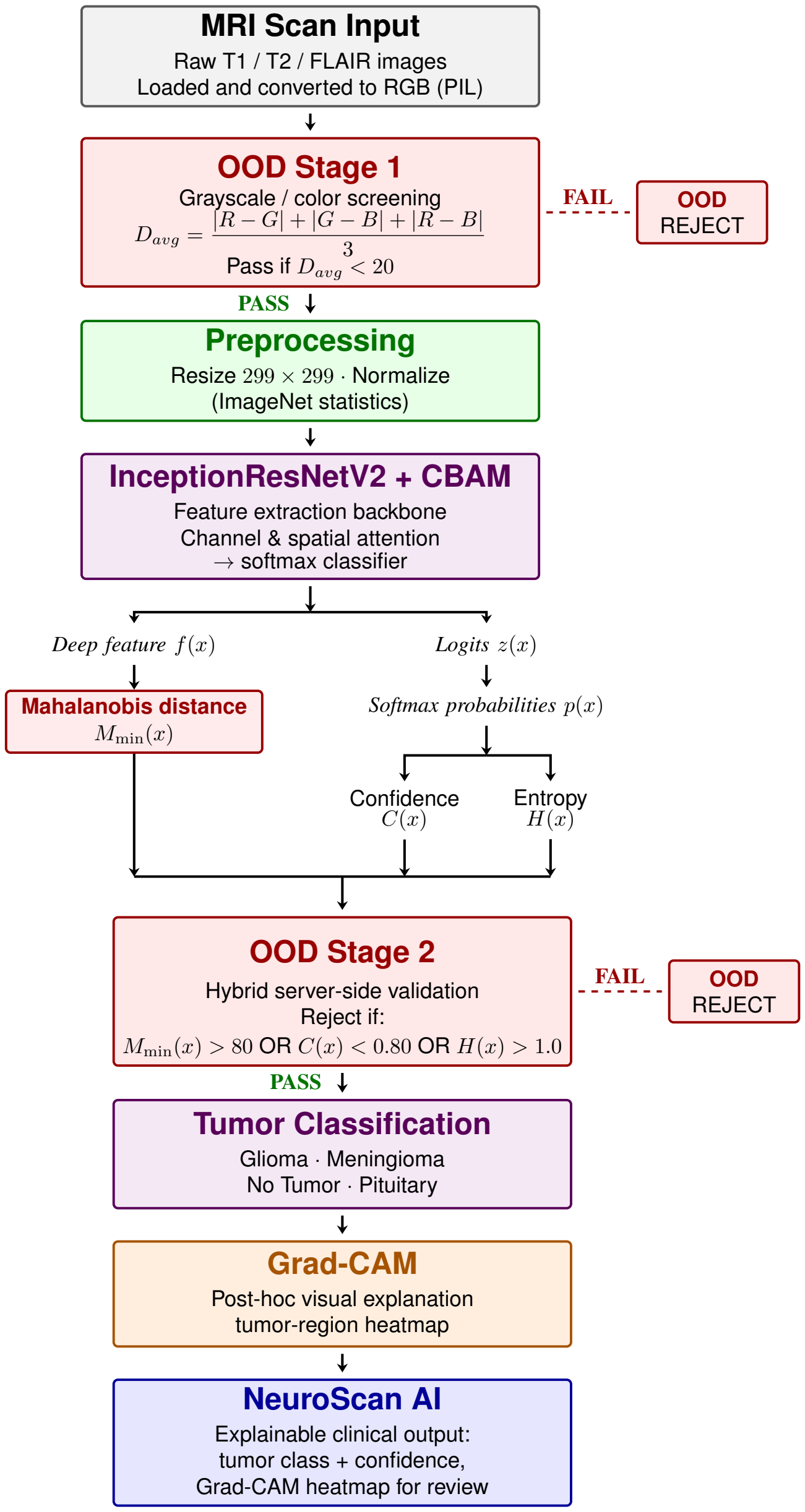

\subsection{Dataset and Preprocessing}
A publicly available Mendeley brain tumor MRI dataset \cite{hira2025mendeley}, available at \url{https://data.mendeley.com/datasets/zwr4ntf94j/1}, comprising 12,064 images across four classes (Table~I) is used. Images are resized to $299\times299$, converted to tensors, and normalized using ImageNet statistics. Training images are augmented with random horizontal flip, rotation ($\pm15^\circ$), affine translation, color jitter, and random erasing; validation/test images are left unaugmented. The source dataset's predefined training (9,650 images) and test (2,414 images) partitions were retained, with 20\% of the training partition reserved for validation. The public release does not provide patient identifiers, so patient-level separation could not be independently verified, a constraint discussed in Section~VI.

\begin{table}[t]
\caption{Dataset Distribution Across Classes}
\centering
\begin{tabular}{lrrrr}
\toprule
\textbf{Class} & \textbf{Train} & \textbf{Test} & \textbf{Total} & \textbf{\%} \\
\midrule
Glioma      & 3{,}018 & 755 & 3{,}773 & 31.27 \\
Meningioma  & 2{,}183 & 546 & 2{,}729 & 22.62 \\
No Tumor    & 1{,}945 & 487 & 2{,}432 & 20.16 \\
Pituitary   & 2{,}504 & 626 & 3{,}130 & 25.95 \\
\midrule
Total       & 9{,}650 & 2{,}414 & 12{,}064 & 100 \\
\bottomrule
\end{tabular}
\end{table}

\subsection{InceptionResNetV2 Backbone}
InceptionResNetV2 \cite{szegedy2017inceptionresnet} combines the multi-scale feature extraction of Inception modules with the training stability of residual connections. The stem reduces the $299\times299$ input to a $35\times35\times256$ feature representation through three convolutional layers and mixed max/average pooling. Five Inception-ResNet-A blocks then process the $35\times35$ feature maps through three parallel convolutional paths of increasing receptive field, fused via a $1\times1$ convolution and combined with a residual connection. A Reduction-A module reduces spatial dimensions to $17\times17$ via concurrent max-pooling and convolutional branches. Ten Inception-ResNet-B blocks then operate at $17\times17$ resolution, using factorized $1\times7$ and $7\times1$ convolutions in place of a single $7\times7$ convolution to reduce parameter count. A Reduction-B module further reduces spatial dimensions to $8\times8$, after which five Inception-ResNet-C blocks apply factorized $1\times3$/$3\times1$ convolutions to produce the final $8\times8\times1536$ feature map used for CBAM refinement and classification. Table~II summarizes the resulting architecture.

\begin{table}[t]
\caption{InceptionResNetV2 Architecture Summary}
\centering
\begin{tabular}{lrrr}
\toprule
\textbf{Block} & \textbf{Output Size} & \textbf{Repetitions} & \textbf{Channels} \\
\midrule
Input               & $299\times299$ & --- & 3 \\
Stem                & $35\times35$   & 1  & 256 \\
Inception-ResNet-A  & $35\times35$   & 5  & 256 \\
Reduction-A         & $17\times17$   & 1  & 896 \\
Inception-ResNet-B  & $17\times17$   & 10 & 896 \\
Reduction-B         & $8\times8$     & 1  & 1792 \\
Inception-ResNet-C  & $8\times8$     & 5  & 1792 \\
Conv2d-7b           & $8\times8$     & 1  & 1536 \\
Global Avg Pool     & $1\times1$     & 1  & 1536 \\
\bottomrule
\end{tabular}
\end{table}

\subsection{CBAM Attention Integration}
CBAM \cite{woo2018cbam} is inserted immediately after the final convolutional layer (\texttt{conv2d\_7b}), refining the 1536-channel feature map before classification. Given an intermediate feature map $F$, CBAM sequentially applies channel and spatial attention:
\begin{equation}
F' = M_c(F) \otimes F, \qquad F'' = M_s(F') \otimes F'
\end{equation}
where
\begin{equation}
M_c(F) = \sigma\big(\mathrm{MLP}(\mathrm{AvgPool}(F)) + \mathrm{MLP}(\mathrm{MaxPool}(F))\big)
\end{equation}
recalibrates channels using a shared multilayer perceptron with reduction ratio $r=16$, and
\begin{equation}
M_s(F') = \sigma\big(f^{7\times7}([\mathrm{AvgPool}(F');\mathrm{MaxPool}(F')])\big)
\end{equation}
applies a $7\times7$ convolution over pooled spatial descriptors to recalibrate spatial locations (Fig.~2). Channel attention emphasizes discriminative texture and intensity patterns associated with tumor tissue; spatial attention localizes tumor-relevant regions while suppressing background and skull structures. CBAM is applied at the final feature map rather than inside intermediate residual blocks: this placement refines the high-level, task-specific representation used for classification while leaving the ImageNet-pretrained low- and mid-level filters undisturbed, a common choice in transfer-learning settings with limited medical training data. A formal ablation quantifying the marginal contribution of CBAM (with vs.\ without) and of multi-stage placement is identified as a priority direction for future work (Section~VI).

\begin{figure}[t]
\centering
\includegraphics[width=0.95\linewidth]{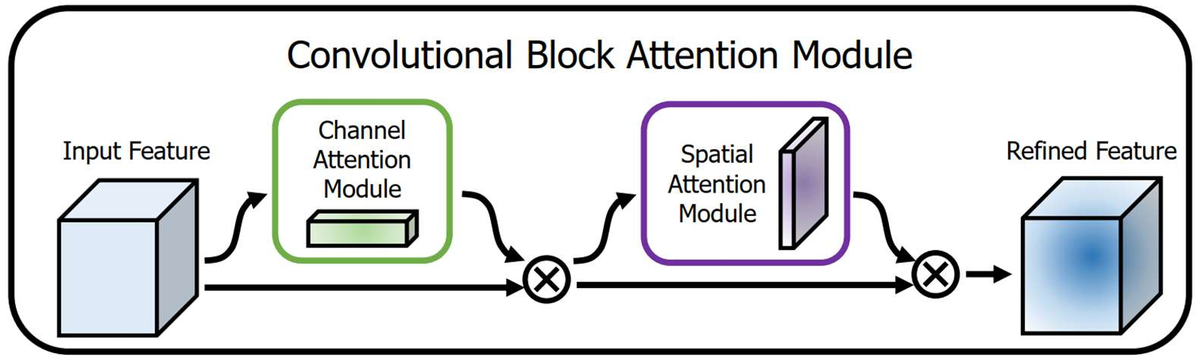}
\caption{CBAM module architecture: sequential channel and spatial attention refinement.}
\label{fig:fig2}
\end{figure}

\subsection{Grad-CAM Explainability}
Grad-CAM \cite{selvaraju2017gradcam} generates a class-discriminative localization map from the target layer's activations $A^k$ and the gradients of the class score $y^c$ with respect to them:
\begin{equation}
\alpha_k^c = \frac{1}{Z}\sum_i\sum_j \frac{\partial y^c}{\partial A_{ij}^k}, \qquad
L^c_{\mathrm{Grad\text{-}CAM}} = \mathrm{ReLU}\Big(\sum_k \alpha_k^c A^k\Big)
\end{equation}
The target layer is \texttt{conv2d\_7b}: late-stage convolutional layers encode high-level semantic content while retaining sufficient spatial resolution for localization. The resulting $8\times8$ map is bilinearly upsampled to $299\times299$ and overlaid on the input MRI, highlighting the regions most responsible for the predicted class. This spatial resolution constrains localization precision, an acknowledged limitation discussed in Section~VI.

\subsection{Two-Stage Out-of-Distribution Detection}
A fast client-side check rejects clearly non-MRI (color) images using the mean inter-channel difference
\begin{equation}
D_{avg} = \frac{|R-G|+|G-B|+|R-B|}{3}, \quad \text{reject if } D_{avg}\geq 20.
\end{equation}
This heuristic targets color (RGB) non-MRI inputs such as photographs; grayscale-modality inputs (e.g., CT, ultrasound) pass this stage unfiltered and rely entirely on the second, feature- and prediction-level stage. Images that pass are validated server-side against three criteria computed from the classifier's own forward pass. The first is Mahalanobis distance over deep feature embeddings $f(x)$:
\begin{equation}
M_c(x) = \sqrt{(f(x)-\mu_c)^T \Sigma^{-1} (f(x)-\mu_c)}
\end{equation}
with class means $\mu_c$ and shared covariance $\Sigma$ estimated from training features, giving a minimum class-conditional distance $M_{min}(x)$. The second is softmax confidence,
\begin{equation}
C(x) = \max_c \; \mathrm{softmax}\big(z(x)\big)_c,
\end{equation}
where $z(x)$ are the classifier logits. The third is predictive entropy over the softmax distribution $p(x)$,
\begin{equation}
H(x) = -\sum_c p_c(x)\log p_c(x).
\end{equation}
An input is flagged OOD if any one of the three criteria is triggered:
\begin{equation}
M_{min}(x) > 80 \;\;\text{or}\;\; C(x) < 0.80 \;\;\text{or}\;\; H(x) > 1.0.
\end{equation}
This OR-combined rule makes Stage~2 a hybrid feature-space and predictive-uncertainty validator rather than a Mahalanobis-only check: an input can be rejected either because its deep features lie far from every class-conditional distribution, or because the classifier itself is under-confident or highly uncertain about it, independent of the Mahalanobis distance. The two-stage design trades some generality (Stage~1 is color-specific; Stage~2 is threshold-based rather than learned) for low latency at the point of care, an explicit design choice discussed further in Section~VI.

\subsection{Training Configuration}
The model is trained end-to-end with the Adam optimizer (initial learning rate $1\times10^{-4}$, StepLR decay with step size 5 and $\gamma=0.5$), batch size 32, cross-entropy loss, weight decay $1\times10^{-4}$, and dropout 0.3 in the classification head, for up to 20 epochs with early stopping (patience 5) on an 80:20 train--validation split. The implementation uses PyTorch and the \texttt{timm} library on a Kaggle NVIDIA T4 GPU; the InceptionResNetV2 backbone is initialized with ImageNet-pretrained weights and fine-tuned end-to-end together with the appended CBAM module and classification head.

\subsection{Deployment: NeuroScan AI}
The trained pipeline is deployed as a full-stack clinical web application, NeuroScan AI, comprising a Python Flask backend and a browser-based frontend. The backend exposes a RESTful endpoint (\texttt{/predict}) that, for each incoming request, performs: (1) grayscale-heuristic OOD screening; (2) standardized image preprocessing; (3) feature extraction via the InceptionResNetV2--CBAM backbone, followed by server-side hybrid OOD/uncertainty validation (Mahalanobis distance, softmax confidence, and predictive entropy) and, for inputs that pass, tumor classification; (4) Grad-CAM heatmap generation; and (5) JSON response construction containing the predicted class, per-class probabilities, and Base64-encoded visualizations. The frontend provides a drag-and-drop upload interface, an animated inference-progress indicator, a four-panel Grad-CAM visualization (original image, heatmap, overlay, and thresholded concentrated activation map), and explicit research-use disclaimers. This deployment demonstrates that the proposed pipeline operates end-to-end in real time rather than only under offline batch evaluation, directly addressing the deployability gap that remains largely unaddressed across the broader brain tumor XAI literature surveyed in Section~II.

\section{Experimental Results}

\subsection{Classification Performance}
Training accuracy improved from 86.33\% to 99.79\% over 20 epochs, while validation accuracy peaked at 98.70\% (epoch 19); test accuracy was 96.73\%. This approximately 2-point validation--test gap is consistent with normal generalization variance for a held-out partition rather than indicative of overfitting, since the reported checkpoint was selected on validation loss rather than test performance. Table~III and Fig.~3(a) report per-class AUC; Pituitary achieved a near-perfect AUC of 1.000, with all other classes exceeding 0.995 and a macro-average AUC of 0.9965.

\begin{table}[t]
\caption{Per-Class AUC Scores}
\centering
\begin{tabular}{lr}
\toprule
\textbf{Class} & \textbf{AUC} \\
\midrule
Glioma      & 0.995 \\
Meningioma  & 0.997 \\
No Tumor    & 0.995 \\
Pituitary   & 1.000 \\
\midrule
Macro-Avg   & 0.9965 \\
Micro-Avg   & 0.9956 \\
\bottomrule
\end{tabular}
\end{table}

\begin{figure}[t]
\centering
\begin{minipage}{0.48\columnwidth}
    \centering
    \includegraphics[width=\linewidth]{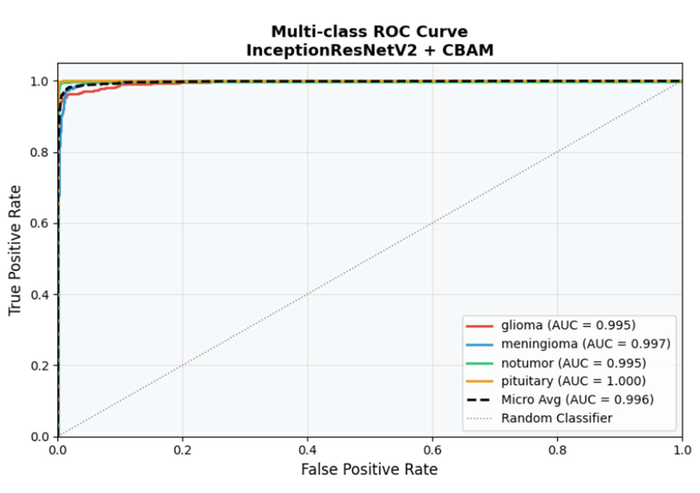}\\[-1mm]
    {\small (a)}
\end{minipage}
\hfill
\begin{minipage}{0.48\columnwidth}
    \centering
    \includegraphics[width=\linewidth]{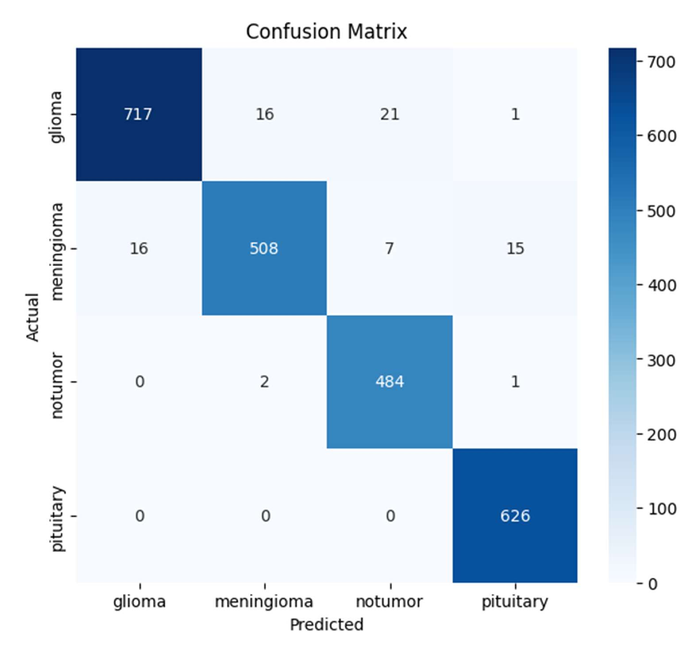}\\[-1mm]
    {\small (b)}
\end{minipage}
\caption{(a) Multi-class ROC curves (One-vs-Rest) for the proposed model. (b) Confusion matrix on the held-out test set.}
\label{fig:fig3}
\end{figure}

\begin{table}[t]
\caption{Grad-CAM Activation Interpretation}
\centering
\begin{tabular}{lll}
\toprule
\textbf{Class} & \textbf{Activation Pattern} & \textbf{Clinical Meaning} \\
\midrule
Glioma      & Asymm.\ intra-parenchymal & Infiltrative tumor \\
Meningioma  & Peripheral extra-axial    & Dural-based lesion \\
Pituitary   & Central basal (sellar)    & Sellar region tumor \\
No Tumor    & Diffuse, low intensity    & Absence of focal lesion \\
\bottomrule
\end{tabular}
\end{table}

\subsection{Confusion Matrix and Failure Analysis}
On the held-out test set (Fig.~3(b)), Pituitary achieved perfect recall (626/626, 100\%), No Tumor 99.38\%, Glioma 94.97\%, and Meningioma 93.04\%. The dominant confusion is between Glioma and Meningioma (16 Glioma slices predicted as Meningioma and 16 Meningioma slices predicted as Glioma). This bidirectional error pattern is clinically plausible: both tumor types can exhibit overlapping intensity distributions, irregular margins, and comparable contrast-enhancement patterns on selected axial slices, particularly for small tumors, atypical locations, or slices captured near the tumor periphery where meningioma's characteristic extra-axial dural attachment is not clearly visible. This bidirectional pattern suggests no systematic class bias and may reflect overlapping imaging characteristics. Notably, the same Glioma--Meningioma confusion pattern is independently reported by Yoon and Lin \cite{yoon2025lime} across VGG16, ResNet50, and EfficientNetB0, suggesting this is a dataset- and class-level difficulty rather than an architecture-specific weakness of either study.

\subsection{Grad-CAM Explainability Across All Tumor Classes}
Grad-CAM activations were examined qualitatively across all four classes to assess correspondence with clinically expected tumor locations (Table~IV, Fig.~4).

\begin{figure*}[!t]
\centering
\begin{minipage}{0.23\textwidth}
    \centering
    \includegraphics[width=\linewidth]{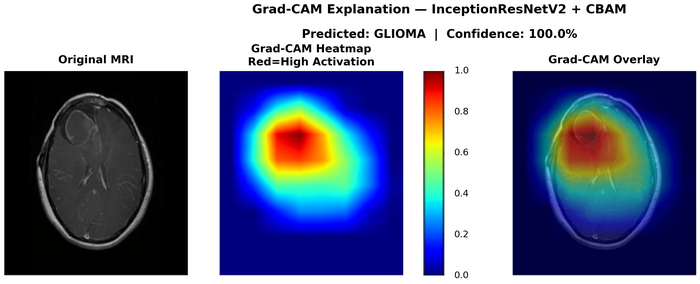}\\[-1mm]
    {\small (a)}
\end{minipage}
\hfill
\begin{minipage}{0.23\textwidth}
    \centering
    \includegraphics[width=\linewidth]{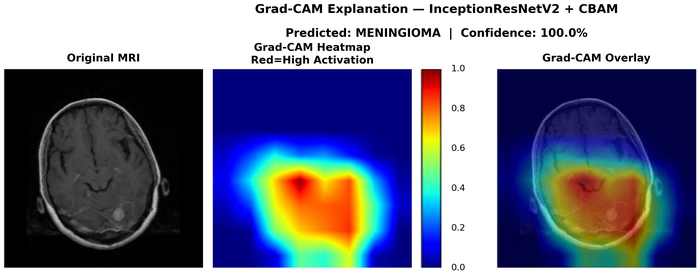}\\[-1mm]
    {\small (b)}
\end{minipage}
\hfill
\begin{minipage}{0.23\textwidth}
    \centering
    \includegraphics[width=\linewidth]{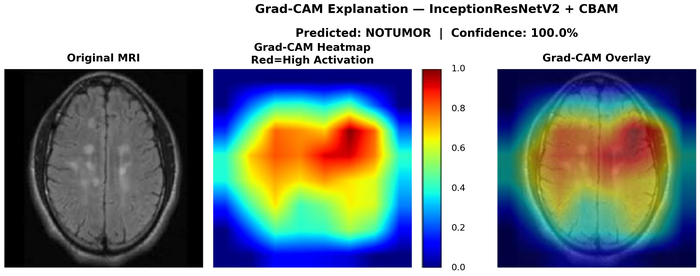}\\[-1mm]
    {\small (c)}
\end{minipage}
\hfill
\begin{minipage}{0.23\textwidth}
    \centering
    \includegraphics[width=\linewidth]{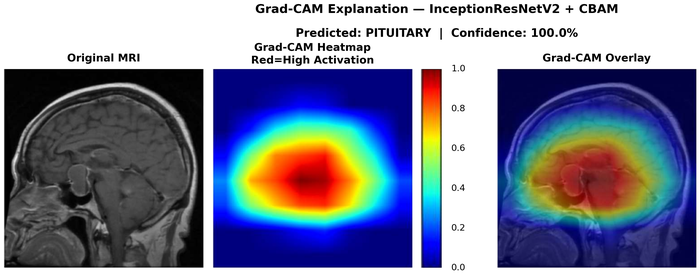}\\[-1mm]
    {\small (d)}
\end{minipage}
\caption{Grad-CAM examples (original, heatmap, overlay) for (a) glioma, (b) meningioma, (c) no-tumor, and (d) pituitary classes.}
\label{fig:fig4}
\end{figure*}

For glioma (Fig.~4(a)), activation concentrates asymmetrically within one cerebral hemisphere, consistent with the lateralized, infiltrative growth pattern of gliomas. For meningioma (Fig.~4(b)), activation localizes near the peripheral, dural-adjacent lesion boundary, consistent with meningioma's characteristic extra-axial origin. For pituitary tumors (Fig.~4(d)), activation concentrates centrally in the sellar/suprasellar region, matching the anatomical location of the pituitary gland. For no-tumor cases (Fig.~4(c)), activation is diffuse and low-intensity across both hemispheres without focal concentration, consistent with the absence of a discriminative lesion. This consistent correspondence across all four classes provides qualitative evidence that the network attends to clinically plausible regions; however, it does not by itself establish explanation faithfulness or exclude reliance on spurious correlations. Quantitative validation is identified as a priority extension in Section~VI.

\subsection{Out-of-Distribution Detection Performance}
The two-stage OOD pipeline was evaluated on a preliminary benchmark of 50 in-distribution MRI images and 50 out-of-distribution images (photographs, X-rays, non-medical images). As shown in Table~V, the combined framework achieves 99\% detection accuracy with a 2\% false-positive rate on this benchmark. This benchmark is limited in scale ($n=100$) and composition: it does not yet include other clinical imaging modalities (e.g., brain CT, chest X-ray, ultrasound) or corrupted/artifact-laden MRI slices, which would constitute a more challenging near-OOD test; expanding the benchmark along these axes, and reporting threshold-independent metrics such as AUROC and FPR95, is identified as priority future work (Section~VI).

\begin{table}[t]
\caption{OOD Detection Performance}
\centering
\begin{tabular}{lrrr}
\toprule
\textbf{Method} & \textbf{TPR} & \textbf{FPR} & \textbf{Acc.} \\
\midrule
Client (grayscale)       & 100.0\% & 6.0\% & 97.0\% \\
Server (Hybrid OOD)      & 98.0\%  & 2.0\% & 98.0\% \\
Combined (two-stage)     & 100.0\% & 2.0\% & 99.0\% \\
\bottomrule
\end{tabular}
\end{table}

\subsection{Qualitative Comparison with Reported Literature}
Table~VI situates the proposed framework against prior CNN-based brain tumor classifiers, including the closely related interpretability study of Yoon and Lin \cite{yoon2025lime}, extending the comparison with explainability (XAI), OOD detection, and deployment status. Accuracy figures are cross-dataset (different sources, splits, and preprocessing) rather than head-to-head under identical protocols; baseline models were not retrained on the present dataset and split, so this table should be read as a positioning summary against reported literature rather than a controlled benchmark. While several baselines, including \cite{yoon2025lime}, report competitive or higher raw accuracy for individual architectures, none combine an attention-modified backbone, OOD-based input validation, and a deployed clinical interface; the proposed framework is the only one in this comparison to provide all three alongside near-state-of-the-art accuracy.

\begin{table}[!t]
\centering
\caption{Qualitative Comparison with Reported Literature}
\label{tab:comparison}
\footnotesize
\setlength{\tabcolsep}{2.2pt}
\renewcommand{\arraystretch}{0.92}
\begin{tabular}{p{0.31\columnwidth} c c c c c}
\toprule
Method & Acc. & Att. & XAI & OOD & Dep. \\
\midrule
CNN (Abiwinanda et al.) \cite{abiwinanda2019cnn} & 84.19\% & No & No & No & No \\
VGG19 \cite{swati2019vgg19} & 94.82\% & No & No & No & No \\
Yoon \& Lin \cite{yoon2025lime} & $\geq$96\% & No & Yes* & No & No \\
Proposed (validation) & 98.70\% & Yes & Yes & Yes & Yes \\
Proposed (test) & 96.73\% & Yes & Yes & Yes & Yes \\
\bottomrule
\end{tabular}
\\[2pt]
\raggedright{\footnotesize *Comparative Grad-CAM/LIME interpretability analysis of unmodified architectures, not integrated into a deployed pipeline.}
\end{table}

\section{Trustworthy AI Discussion}
The results demonstrate that attention-guided classification, post-hoc
explainability, and OOD-aware validation can be combined within a
single deployed framework without sacrificing classification
performance. The resulting system provides accurate predictions,
visual explanations through Grad-CAM, and automated screening for
input validity.

\section{Threats to Validity}
The public dataset lacks patient identifiers, so patient-wise separation cannot be independently verified. Grad-CAM is qualitative; future validation should include faithfulness metrics (e.g., Deletion/Insertion AUC, confidence drop under top-region masking) and overlap with radiologist-annotated regions. The preliminary OOD benchmark ($n=100$) excludes harder near-OOD inputs, and Stage-2 thresholds were selected on validation data rather than calibrated to predefined operating points. Table~VI is cross-study positioning, not a controlled same-split benchmark. These limitations bound the trustworthiness claims but not the demonstrated system-level integration; comprehensive validation remains future work. Priorities are patient-level external validation, quantitative XAI, broader near-OOD testing with AUROC/FPR95, controlled same-split baseline retraining, and InceptionResNetV2 CBAM ablations of attention removal and placement.

\section{Conclusion and Future Work}
This paper presented a trustworthy-AI-oriented framework for brain tumor MRI classification that integrates an attention-enhanced InceptionResNetV2-CBAM classifier, Grad-CAM explainability across all four tumor classes, a two-stage hybrid OOD detection pipeline combining grayscale screening with Mahalanobis distance, softmax confidence, and predictive entropy, and a deployed clinical web application, NeuroScan AI. The framework achieves a macro-average AUC of 0.9965, 96.73\% test accuracy, and 99\% detection accuracy on a preliminary OOD benchmark. The primary contribution is the integration of attention, OOD screening, and real-time deployment rather than superior raw classification accuracy; validation priorities are detailed in Section~VI. Taken together, the proposed framework demonstrates that diagnostic accuracy, interpretable predictions, input-validation robustness, and deployable clinical workflows can be integrated within a single deep learning system, while further evaluation is needed before clinical deployment.

\end{document}